\documentclass[10pt,twocolumn,letterpaper]{article}

\usepackage{cvpr}
\usepackage{times}
\usepackage{epsfig}
\usepackage{graphicx}
\usepackage{amsmath}
\usepackage{amssymb}
\usepackage[font={small}]{caption}

\usepackage{url}            
\usepackage{booktabs}       
\usepackage{nicefrac}       
\usepackage{microtype}      
\usepackage[inline]{enumitem}
\usepackage{color}

\usepackage{makecell}

\newcommand{\beas}{\begin{eqnarray*}}
	\newcommand{\eeas}{\end{eqnarray*}}
\newcommand{\bea}{\begin{eqnarray}}
\newcommand{\eea}{\end{eqnarray}}
\newcommand{\bes}{\begin{equation*}}
\newcommand{\ees}{\end{equation*}}
\newcommand{\be}{\begin{equation}}
\newcommand{\ee}{\end{equation}}

\newcommand{\cY}{{\cal Y}}

\newcommand{\norm}[1]{\left\lVert#1\right\rVert}

\newcommand{\idan}[1]{{\color{red}{Idan: #1}}}

\makeatletter
\usepackage{xspace}
\def\@onedot{\ifx\@let@token.\else.\null\fi\xspace}
\DeclareRobustCommand\onedot{\futurelet\@let@token\@onedot}
\newcommand{\figref}[1]{Fig\onedot~\ref{#1}}

\newcommand{\secref}[1]{Sec\onedot~\ref{#1}}
\newcommand{\tabref}[1]{Tab\onedot~\ref{#1}}

\def\eg{\emph{e.g}\onedot} 
\def\ie{\emph{i.e}\onedot} 
\def\cf{\emph{cf}\onedot} 
 \def\vs{\emph{vs}\onedot}
 
\def\etal{\emph{et al}\onedot}

\usepackage[pagebackref=true,breaklinks=true,letterpaper=true,colorlinks,bookmarks=false]{hyperref}

\cvprfinalcopy 

\def\cvprPaperID{6902} 

\ifcvprfinal\fi

\begin{document}

\title{A Simple Baseline for Audio-Visual Scene-Aware Dialog\vspace{-0.5cm}}

\author{
Idan Schwartz\textsuperscript{1}, Alexander Schwing\textsuperscript{2}, Tamir Hazan\textsuperscript{1}\\
\textsuperscript{1}Technion \hspace{1cm}\textsuperscript{2}UIUC\\
{\tt\footnotesize idanschwartz@gmail.com, aschwing@illinois.edu, tamir.hazan@technion.ac.il}
\vspace{-0.5cm}
}


\maketitle
\begin{abstract}
The recently proposed audio-visual scene-aware dialog task paves the way to a more data-driven way of learning virtual assistants, smart speakers and car navigation systems.  However, very little is known to date about how to effectively extract meaningful information from a plethora of sensors that pound the computational engine of those devices. Therefore, in this paper, we provide and carefully analyze a simple baseline for audio-visual scene-aware dialog which is trained end-to-end. Our method differentiates in a data-driven manner useful signals from distracting ones using an attention mechanism. We evaluate the proposed approach on the recently introduced and challenging audio-visual scene-aware dataset, 
and demonstrate the key features that permit to outperform the current state-of-the-art by more than 20\% on CIDEr. 
\end{abstract}
\section{Introduction}

We are interacting with a dynamic environment which constantly stimulates our brain  via visual and auditory signals. Despite the huge amount of different information that is permanently occupying our nervous system, we are often easily able to quickly discern important cues from data that is irrelevant. Telling apart useful information from distracting aspects is also an important ability for virtual assistants, car navigation systems, or smart speakers.  However present day technology uses a chain of components from speech recognition and dialog management to sentence generation and speech synthesis, making it hard to design a holistic and entirely data-driven approach. 

For instance, in computer vision, a tremendous amount of recent work has focused on image captioning~\cite{VinyalsCVPR2015,johnson2016densecap,ChenCVPR2015,donahue2015long,show_tell,MaoARXIV2014,you2016image,KarpathyCVPR2015,WangNIPS2017,AnejaCVPR2018,DeshpandeARXIV2018,ChatterjeeECCV2018}, visual question generation~\cite{LiARXIV2017DualVQAVQG,VQG,mostafazadeh2017image,JainCVPR2017}, visual question answering~\cite{AnatolICCV2015,GaoNIPS2015,ShihCVPR2016,rennips2015,MalinowskiICCV2015,XiongICML2016,XuARXIV2015,YangCVPR2016,SchwartzNIPS2017,SchwartzCVPR2019,NarasimhanNIPS2018,NarasimhanECCV2018}, and very recently visual dialog~\cite{visdial, visdial_rl,JainCVPR2018,MassicetiARXIV2018}. While those meticulously engineered algorithms have shown promising results in their specific domain, little is known about the end-to-end performance of an entire system. This is partly due to the fact that little data is publicly available to design such an end-to-end algorithm. 

\begin{figure}[t]
\centering
\includegraphics[width=1\linewidth]{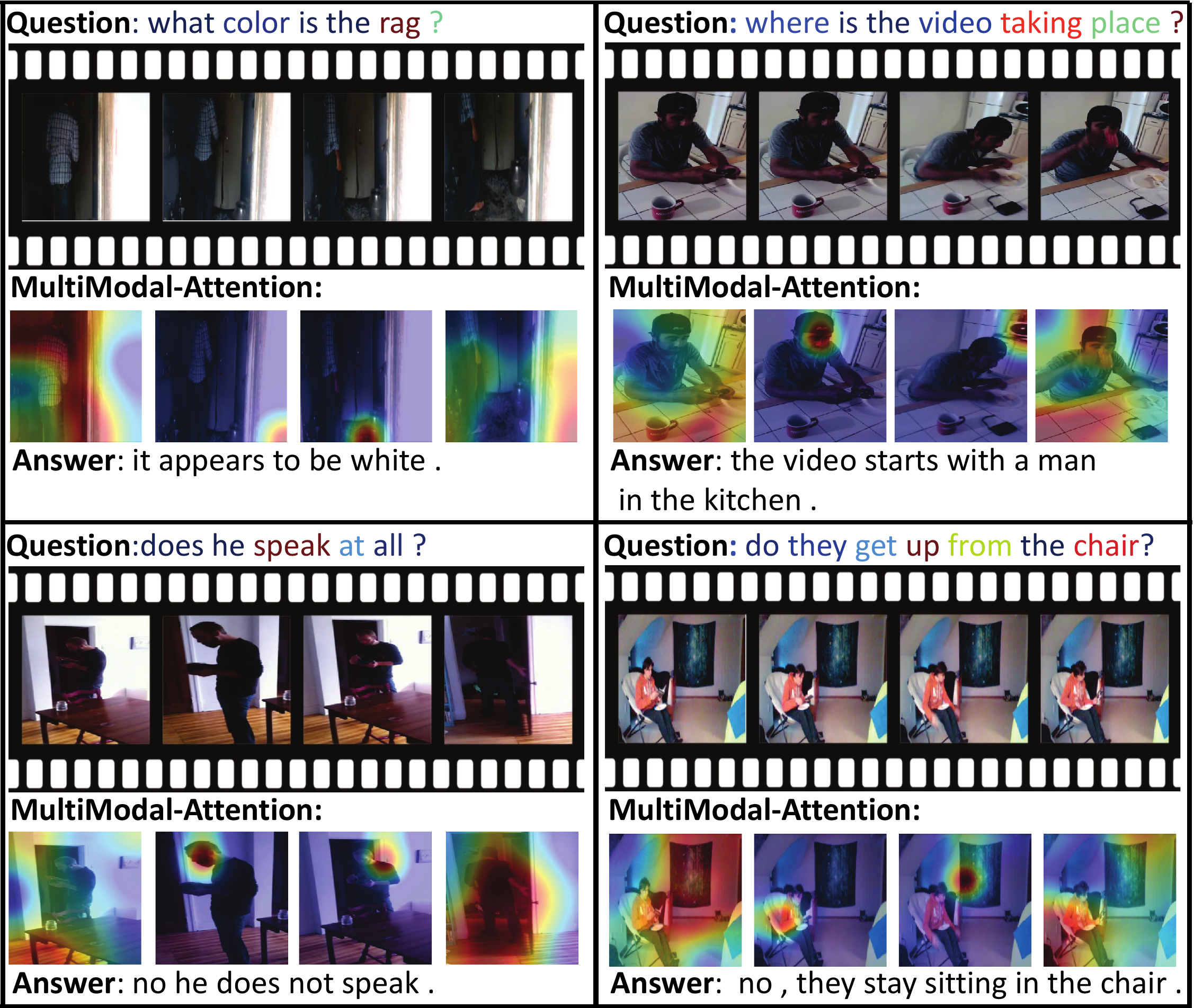}
\vspace{-0.7cm}
\caption{\small We present 4 different questions and the generated answer. Our attention unit is illustrated as well. Our model samples 4 frames, and attends to each frame separately, along with the question and the audio. We observe attention for each frame  to differ, where first and fourth frames are widespread, while the second and third are more specific. Also, the question attention attends to relevant words. We also include the audio modality as input to the attention computation.   }
\label{fig:teaser}
\vspace{-0.7cm}
\end{figure}

Recent work on audio-visual scene aware dialog~\cite{AlamriARXIV2018,HoriARXIV2018} partly addresses this shortcoming and proposes a novel dataset. Different from classical datasets like MSCOCO~\cite{lin2014microsoft}, VQA~\cite{AnatolICCV2015} or  Visual Dialog~\cite{visdial}, this new dataset contains  short video clips, the corresponding audio stream and a sequence of question-answer pairs. While development of an end-to-end data driven system isn't feasible just yet due to the missing speech signal, the new audio-visual scene aware dialog dataset at least permits to develop a holistic dialog management and sentence generation approach taking audio and video signals into account. 

In recent work~\cite{AlamriARXIV2018,HoriARXIV2018}, a baseline for a system based on audio, video and language data was proposed. Compelling results were achieved, demonstrating accurate question answering. The authors demonstrate that multimodal features based on I3D-Kinetics (RGB+Flow)~\cite{carreira2017quo} refined via a carefully designed attention-based mechanism improve the quality of the generated dialog.

However, since much effort was dedicated to collecting the dataset, little analysis of such a holistic system was provided. Moreover, due to tremendous amounts of available data (certainly a ten-fold increase compared to classical visual dialog data) this is by no means trivial. To provide this missing information and to share some insights with the community about how and where to improve, in this paper, we follow the spirit of~\cite{JabriARXIV2016} and demonstrate  
(1) that simply using the question as a signal already permits to outperform the current state-of-the-art; 
(2) that it is crucial to maintain spatial features for the video signal (either VGG19~\cite{Simonyan14c} or I3D-Kinetics~\cite{carreira2017quo}). Reducing every video frame into a single representation drops performance significantly; 
(3) that temporally subsampling the video frames improves the accuracy; 
(4) that using attention over all available data (including different frames) is beneficial. To this end we analyze how to fuse the attended vectors for different data modalities. 

Our simple baseline, which consists of three jointly trained components (data representation extraction, attention and answer generation)  outperforms state-of-the-art by a large margin of 20\% on CIDEr. Improvements of the proposed approach are largely due to the aforementioned four points. 
Results of generated answers are contrasted to the current state-of-the-art in \figref{fig:teaser}. We observe  plausible answers to many questions and attention that focuses on important parts in both video and text. 

\section{Related Work}
A significant amount of research has been conducted regarding image captioning, visual question generation, visual question answering, visual dialog, video data, audio data and multimodal attention models. We briefly review those related areas in the following. 

\noindent\textbf{Image Captioning:} Originally image captioning was formulated as a retrieval problem. The best fitting caption from a set of considered options was found by matching features obtained from the available textual descriptions and the given image. Importantly, the matching function is typically learned using a dataset of  image-caption pairs. While such a formulation permits end-to-end training, assessing the fit of image descriptors to a  large pool of captions is computationally expensive. Moreover, it's likely prohibitive to construct a database of captions that is sufficient for describing even a modestly  large fraction of plausible images. 

To address this challenge, recurrent neural nets (RNNs)  decompose captions into a product space of individual words. 
This technique has recently found widespread use for image captioning because  remarkable results have been demonstrated which are, despite being constructed word by word, syntactically correct most of the time.  
For instance, a CNN to extract image features and a 
language RNN that shares a joint embedding layer was trained~\cite{MaoARXIV2014}. Joint training of a 
 CNN with a language RNN to generate sentences one word at a time was demonstrated in~\cite{show_tell}, and subsequently 
extended~\cite{show_tell} using additional attention parameters which  identify 
salient objects for caption generation.A bi-directional RNN was employed 
along with a structured loss function in a shared vision-language space~\cite{KarpathyCVPR2015}. Diversity was considered, \eg,  by Wang \etal~\cite{WangNIPS2017} and Deshpande \etal~\cite{DeshpandeARXIV2018}. 

\begin{figure*}
	\centering
	\includegraphics[width=1\linewidth]{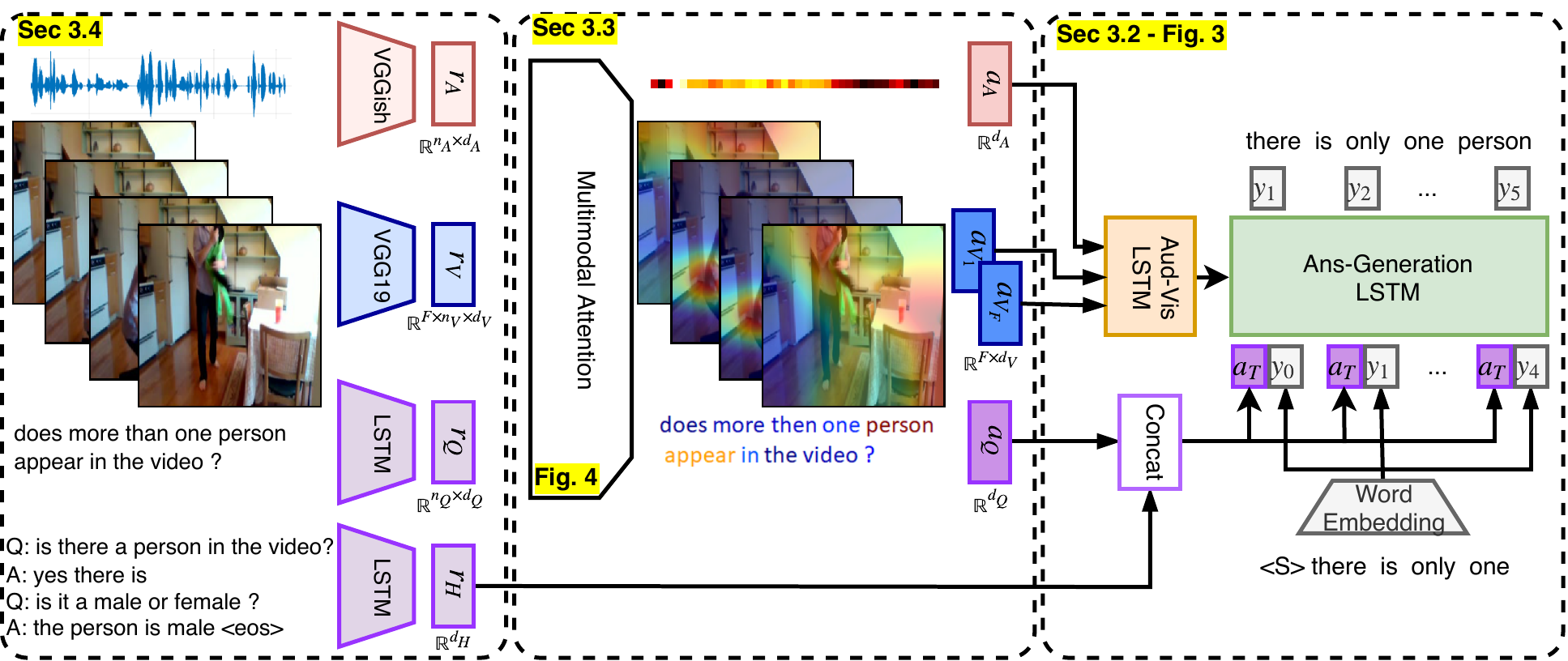}
	\vspace{-0.7cm}
	\caption{\small Overview of our approach for the AVSD task. More details can be found in \secref{sec:AVD}.}
	\label{fig:overview}
	\vspace{-0.5cm}
\end{figure*}

\noindent\textbf{Visual Question Answering:} Beyond generating a caption for an image, a large amount of work has focused on answering a question about a given image. 
On a plethora of datasets~\cite{MalinowskiNIPS2014, rennips2015,AnatolICCV2015,GaoNIPS2015,ZhuCVPR2016, JohnsonCVPR2017Clevr}, models with multi-modal attention~\cite{lu2016hierarchical,YangCVPR2016,AndreasCVPR2016,DasARXIV2016,FukuiARXIV2016,ShihCVPR2016,XuARXIV2015,SchwartzNIPS2017,SchwartzCVPR2019}, deep net architecture developments~\cite{BenyounesICCV2017Mutan,MalinowskiICCV2015, MaARXIV2015} and  memory nets~\cite{XiongICML2016} have been investigated. 

\noindent\textbf{Visual Question Generation:} In spirit similar to  question answering 
is the task of visual question generation, which is still very much an open-ended topic. For example, Ren \etal~\cite{rennips2015} discuss a rule-based method, converting a given sentence into a corresponding question which has a single word answer. 
Mostafazadeh \etal~\cite{VQG} learned a question generation model with human-authored questions rather than machine-generated descriptions. 
Vijayakumar \etal~\cite{VijayakumarARXIV2016} have shown  results for this task as well. Different from the two aforementioned techniques, Jain \etal~\cite{JainCVPR2017} argued for more diverse predictions and use a variational auto-encoder approach. Li \etal~\cite{LiARXIV2017DualVQAVQG} discuss VQA and VQG as dual tasks and suggest a joint training. They take advantage of the state-of-the art VQA model by Ben-younes~\etal~\cite{BenyounesICCV2017Mutan} and report improvements for both VQA and VQG. 

\noindent\textbf{Visual Dialog:} Visual dialog~\cite{visdial} combines the three aforementioned tasks. Strictly speaking it requires both generation of questions and corresponding answers. Originally, visual dialog required  to only predict the answer for a given question, a given image and a provided history of question-answer pairs. While this  resembles the VQA task, different approaches, \eg, also based on reinforcement learning, 
have been proposed recently~\cite{kottur2018visual, visdial_rl, JainCVPR2018,MassicetiARXIV2018, wu2018you}.

\noindent\textbf{Video Data:} A variety of tasks like video paragraph captioning~\cite{yu2016video}, video object segmentation~\cite{Perazzi2016}, pose estimation~\cite{ZhangICCV2015}, video classification~\cite{KarpathyCVPR2014}, and action recognition~\cite{SimonyanNIPS2014} have used video data for a long time. Probably most related to our approach are video classification and action recognition since both techniques also extract a representation from a video. While the extracted representation is subsequently used for either classification or action recognition, we employ the representation to more accurately answer a question. Commonly used feature representations for either video classification or action recognition are  I3D-based features by Carreira \etal\cite{carreira2017quo},  extracted from an action recognition dataset. With proper fine-tuning the I3D-based features proved to be better than the classical approaches, such as C3D~\cite{tran2015learning} that capture spatiotemporal information via a 3D CNN. In this work, we assess a na\"ive feature extractor based on VGG~\cite{Simonyan14c}, and demonstrate that for video-reasoning, careful reduction of the spatial dimension is more crucial than the type of extracted features used to embed the video frames. Wang \etal~\cite{wang2016temporal} 
showed that working with video frame samples, achieves not only efficiency, but also improves performance compared to a conservative dense temporal representation. 
Recently, Zhou \etal~\cite{zhou2017temporal} further extended those ideas, and suggested to capture relational temporal relationships between the sampled frames, relying on the relational-networks concept~\cite{santoro2017simple}. We follow those ideas by also sub-sampling a small set of frames uniformly. Our model further advances those concepts, by exploiting spatial relationships between sampled temporal frames via a high-order multimodal attention module, where each video frame is treated as a separate modality. Li \etal~\cite{li2018videolstm} propose the Video-LSTM model, which uses attention to emphasis relevant locations, during LSTM video encoding. Our approach differs in that attention on one frame can influence attention on other frames which isn't the case in their model. 

\noindent\textbf{Audio Data:} Audio data   gained popularity in the vision community  recently. For instance, prediction of pose given audio input~\cite{ShlizermanCVPR2018}, learning of audio-visual object models from unlabeled video for audio source separation in novel videos~\cite{GaoCVPRW2018,OwensECCV2018}, use of video and audio data for acoustic scene/object classification~\cite{aytar2016soundnet}, source separation was also considered in~\cite{EphratSIGGRAPH2018} and  learning to see using audio~\cite{OwensIJCV2018}. 

\noindent\textbf{Multimodal Attention:} Multimodal attention has been a prominent component in  tasks which operate on different input data. Xu \etal~\cite{show_tell} showed an encoder decoder attention model for image captioning, which was extended  to visual question answering~\cite{XuARXIV2015}. Yang \etal~\cite{YangCVPR2016} propose a multi-step reasoning system using an attention model. Multimodal pooling methods were also explored~\cite{FukuiARXIV2016,kim2016hadamard}. Lu
\etal~\cite{lu2016hierarchical} suggest to produce co-attention for the image and question separately, using a hierarchical and parallel formulation. Schwartz \etal~\cite{SchwartzNIPS2017,SchwartzCVPR2019} later extend this approach to high-order attention applied over image, question and answer modalities via potentials. Similarly, in the visual dialog task, co-attention models have held the state-of-the-art~\cite{wu2017you, lu2017best}  attending over image, question and history in hierarchical manner. 
For audio-visual scene-aware dialog,~\cite{HoriARXIV2018} also use a sum-pooling type of attention, using the question feature along with audio and video modalities separately. In contrast, here we compute attention over each modality via local and cross data evidence, letting all the modalities  interact with each other. 

\section{Audio Visual Scene-Aware Dialog Baselines}
\label{sec:AVD}


Our method has three building blocks: answer generation, attention and data representation as shown in \figref{fig:overview}. 



\subsection{Answer Generation}
\label{sec:gen}
\begin{figure}
	\centering
	\includegraphics[width=1\linewidth]{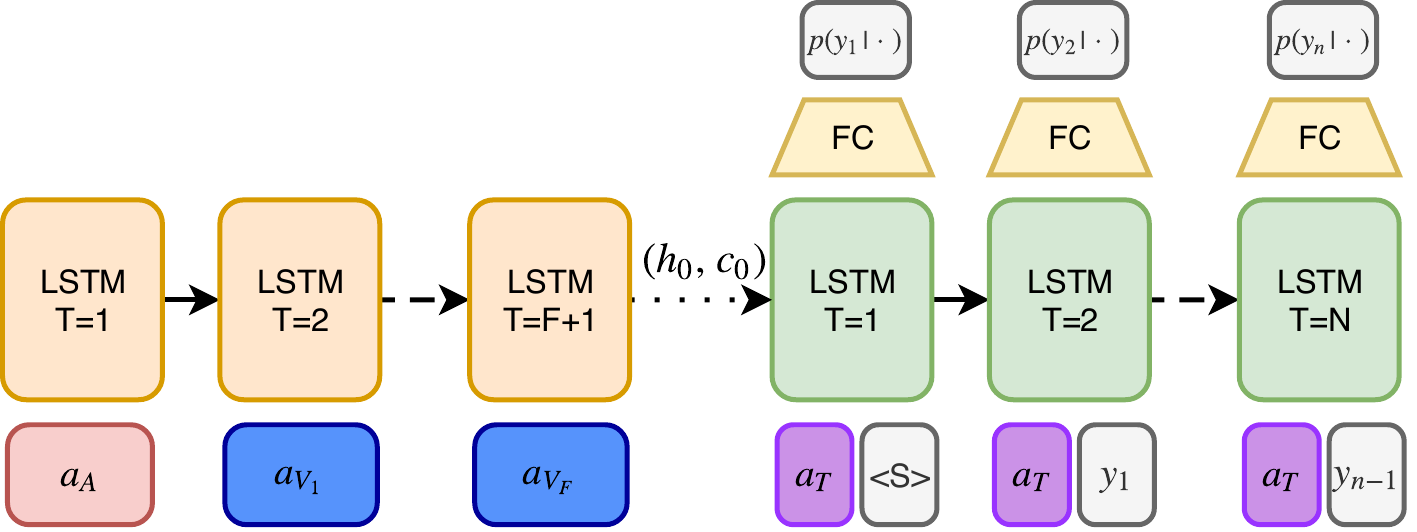}
	\vspace{-0.7cm}
   \caption{\small Our decoder for audio-visual scene-aware dialog. We start with encoding of attended audio and video vectors using the Aud-Vis LSTM (orange colored), followed by the Ans-Generation LSTM that receives the textual data concatenated with the previous answer word (green colored).
   }
\label{fig:answer}
\vspace{-0.7cm}
\end{figure}

We are interested in predicting an answer $y = (y_1, \ldots, y_n)$ consisting of $n$ words $y_i\in\cY_i = \{1, \ldots, |\cY_i|\}$ each arising from a vocabulary of possible words $\cY_i$. Given data $x = (Q, V, A, H)$ which subsumes, a question $Q$, a subsampled video $V = (V_1, \ldots, V_{F})$ composed of $F$ frames, the corresponding audio signal $A$, and a history of past question-answer pairs $H$, we construct a probability model over the set of possible words for the answer generation task. To this end, we formulate prediction of the answer as inference in a recurrent model where the joint probability is given by the product of conditionals, \ie,  
$$
p(y|x) = \prod_{i=1}^n p(y_i|y_{<i},x).
$$
Note that, for now, we condition on all the data $x$ for readability and provide details later. Instead of conditioning the probability of the current word $p(y_i|y_{<i},x)$ on its entire past $y_{<i}$, we combine two recurrent nets: an audio-visual recurrent net that generates the temporal information which is fed as an initialization to the answer generating recurrent net. See \figref{fig:answer} for a schematic.

\noindent{\bf Audio-visual LSTM-net:} It operates on an attended audio embedding $a_A$ and attended video embeddings $a_{V_1},...,a_{V_F}$ for each of the $F$ frames $f \in \{1, \ldots, F\}$. This LSTM-net has $F+1$ units, the first unit's input is the attended audio vector, and the input to the $F$ subsequent units are the attended video representations $a_{V_1},\ldots,a_{V_F}$. The context vector that is generated from this LSTM, \ie, $(h_0, c_0)$ summarizes the audio-visual attention and is provided as input to the answer generation LSTM-net.    

\noindent{\bf Answer generation LSTM-net:} It computes conditional probabilities for the possible words $y_i\in\cY_i$ of the answer $y = (y_1, \ldots, y_n)$. This probability considers the last word and captures context via a  representation $h_{i-1}$ obtained from the previous time-step. 
$$
p(y_i|y_{i-1}, h_{i-1},x) = g_w(y_i, y_{i-1}, h_{i-1},x).
$$
We illustrate the LSTM-net $g_w$ in \figref{fig:answer}. 
Using the initial state $(h_0, c_0)$, the LSTM-net $g_w$ predicts in its $i$-th step a probability distribution $p(y_i| y_{i-1}, h_{i-1}, x)$ 
over words $y_i\in\cY_i$ using as input $y_{i-1}$ and the textual attention vector $a_T = (a_Q, r_H)$: the attended textual vector is a concatenation of the attended question vector $a_Q$ and the  history vector $r_H$, which represents information about question and history data. The output of the LSTM-net is transformed via a FC-layer with a dropout and a softmax to obtain the probability distribution $p(y_i | y_{i-1}, h_{i-1}, x)$. 

\begin{figure}
\centering
\includegraphics[width=1\linewidth]{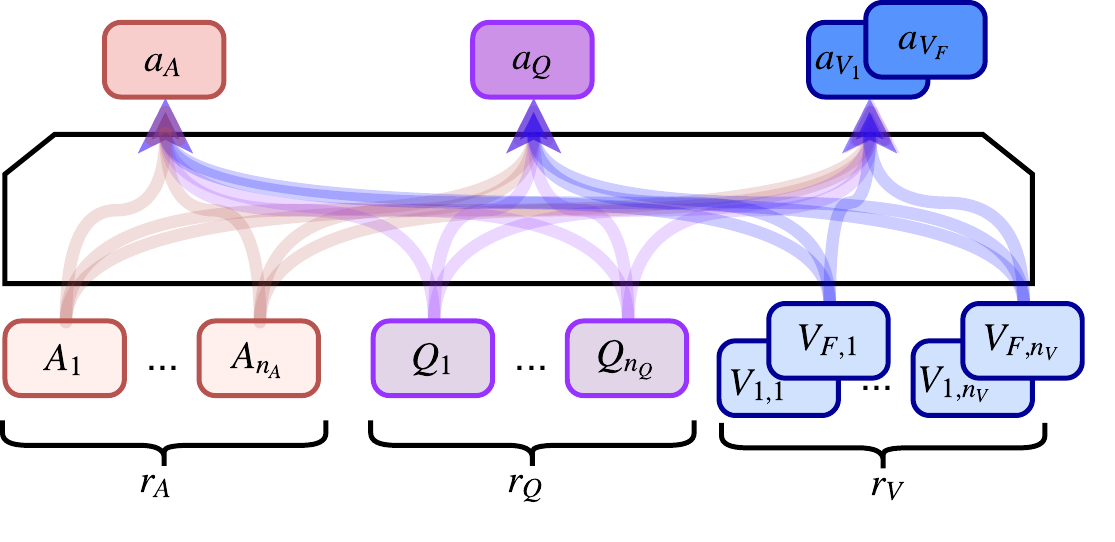}
\vspace{-0.9cm}
\caption{\small Multimodal Attention model for audio-visual scene-aware dialog. We treat each frame as a modality, along with audio and question modality, to total of 6 modalities. Each element attention score is affected not only from local evidence, but also via cross-data interactions of all other elements.}
\label{fig:attention}
\vspace{-0.5cm}
\end{figure}

\subsection{Attention}
\label{sec:attention}
The attention step provides  an attended representation for the data components, \ie, $ a_{V_f} \in \mathbb{R}^{d_V} $ for frame $f\in\{1, \ldots,F\}$  of the video data, $ a_A \in \mathbb{R}^{d_A} $ for the audio data, and $ a_T \in \mathbb{R}^{d_T} $ for the textual data. These attended representations are obtained by transforming the representations extracted from the raw data, \ie, $r_{V_f} \in \mathbb{R}^{n_V \times d_V}$ for the video data, $r_A\in\mathbb{R}^{n_A \times d_A}$ for the audio data, and for the textual data, $r_Q\in\mathbb{R}^{n_Q \times d_Q}$ as well as $r_H\in\mathbb{R}^{d_H}$ 
which capture signals from the question and history respectively. 
We outline the general procedure in \figref{fig:attention}. 


Formally, we obtain the attended representation
$$
a_\alpha = \sum_{k=1}^{n_\alpha} \alpha_{k} p_\alpha(k),
$$
where $\alpha\in \{\emph{A}, \emph{Q}, \emph{V}_1, \ldots, \emph{V}_F\}$ is used to index the available data components (audio, question, visual frames), $n_\alpha$ is the number of entities in a data component (\eg, the number of words in a question), and $p_\alpha(k) \geq 0$ $\forall \alpha$ is a probability distribution ($\sum_{k=1}^{n_\alpha} p_\alpha(k) = 1$ $\forall \alpha$) over the $n_\alpha$ entity representations of data $\alpha$. For instance, if we let $\alpha = A$ we obtain the attended audio representation $a_A = \sum_{k=1}^{n_A} A_k p_A(k)$. 

We compute the attention via a factor graph attention approach~\cite{SchwartzNIPS2017,SchwartzCVPR2019}. The attention probability distribution over a data source $\alpha$ consists of a log-prior distribution $\pi_\alpha$, a local evidence $l_\alpha$ that relies solely on its data representation $r_\alpha$ and a cross data evidence $c_\alpha$ that accounts for correlations between the  different data representations  $r_\alpha, r_\beta$, for $\beta \in \{\emph{A}, \emph{Q}, \emph{V}_1, \ldots, \emph{V}_F\}$. This probability distribution takes the form:
\beas
p_\alpha(k) \propto \exp\left(\hat w_\alpha \pi_\alpha(k) + l_\alpha(k)
 + c_\alpha(k) \right).
\eeas
The local evidence is $l_\alpha(k) = w_\alpha \left( v_\alpha^\top  \operatorname{relu}(V_\alpha \alpha_k)\right)$, the log-prior is $\pi_\alpha(k)$ and the cross data evidence is
$$
c_\alpha(k) = \sum_{\beta \in {\cal D}}  \frac{w_{\alpha,\beta}}{n_\beta}\sum_{j=1}^{n_\beta} \left( \left(\frac{L_{\alpha} \alpha_k}{\norm{L_{\alpha}  \alpha_k}}\right)^\top \left(\frac{R_{\beta} \beta_j}{\norm{R_{\beta} \beta_j}}\right) \right).
$$
The set ${\cal D}  = \{\emph{A}, \emph{Q}, \emph{V}_1, \ldots, \emph{V}_F\} $ consists of the possible data types.
The trainable parameters of the model are: (1) $V_\alpha, L_\alpha, R_\alpha$ which re-embed the data representation to tune the attention; (2) $v_\alpha$ which scores the local modality; and (3) $\hat w_\alpha, w_\alpha, w_{\alpha,\beta}$ which weight the three components with respect to each other.    

We found the use of attention for history to not yield improvements. Therefore, we obtain the attended textual representation $a_T \in\mathbb{R}^{d_T}$ by concatenating the attended question representation $a_Q\in\mathbb{R}^{d_Q}$ with  the history representation $r_H\in\mathbb{R}^{d_H}$. Consequently, $d_T = d_Q + d_H$. 

\subsection{Data Representation}
\label{sec:data}
The proposed approach relies on representations $r_\alpha$ obtained for a variety of data components which we briefly discuss subsequently. 

\noindent\textbf{Video:} Containing both temporal and spatial information, video data is among the most memory consuming. 
Common practice is to  reduce the spatial information while maintaining attention over the temporal dimension. Instead, we first reduce the temporal dimension, maintaining the ability for spatial attention to reason about the video content. 
To ensure fast training, we reduce  the temporal dimension by sampling $ F $ frames uniformly. For each sampled frame we extract a representation from a deep net trained on ImageNet (in our case VGG19). We then fine tune the representation of each frame  using a 1D conv layer with a bias term. This conv layer is identical for all the $F$ frames. Consequently, we obtain the video representation $ r_V \in \mathbb{R}^{F\times n_V \times d_V} $, where $ F $ is the number of sampled frames, $ n_V $ is the spatial dimension and $ d_V $ is the embedding dimension. 

\noindent\textbf{Audio:} For audio, we extracted features from a strong audio classification model (\ie, VGGish~\cite{hershey2017cnn}) by taking the last representation before the final FC-layer. 
This representation has adaptive temporal length. For each batch we find the maximal temporal length of the audio signal, and zero-padded the shorter audio representations. We then fine-tune each audio file using a 1D conv layer with a bias.  We obtain the audio representation $ r_A \in \mathbb{R}^{n_A \times d_A} $, where $ n_A $ is the maximal temporal length of a given batch and $ d_A $ is the embedding dimension. 

\noindent\textbf{Question:} 
We start with an adaptive-length list of 1-hot word-representations. For each batch we find the longest sentence, and zero-pad  shorter ones. We embed each word using a linear-embedding layer, followed by a single layer LSTM-net with dropout. The last hidden state of the LSTM is the question representation $ r_Q \in \mathbb{R}^{n_Q \times d_Q} $, where $ n_Q $ is the length of the maximal sentence for the given batch and $ d_Q $ is the embedding dimension.

\noindent\textbf{History:} The history data source consists of the past $T$ question-answer pairs, which we denote by $H = (Q,A)_{t\in\{1,\ldots,T\}}$. The history embedding consists of two components: we first embed each question-answer pair $(Q,A)_t$ using a LSTM-net to get $T$ representations of the history. We then feed these representations into another LSTM-net to obtain the vector representation $ r_H \in \mathbb{R}^{d_H} $, where $ d_H $ is the history embedding dimension. 

We embed each question-answer pair $(Q,A)_t$ following the question embedding above. A question-answer pair starts with a list of 1-hot word-representations of the words in the question followed by 1-hot word-representations of the words in the answer. For each batch we find the longest question-answer sequence, and zero-pad the shorter ones. We embed each 1-hot vector using a linear-embedding layer, followed by a two layer LSTM-net with a dropout. The last hidden state of this LSTM-net is the vector representation of $(Q,A)_t$, which we denote by $r_t$. 

We embed the history by feeding $r_1,\ldots,r_T$ to a one layer LSTM-net with  dropout, in order to capture the temporal aspect of the question-answer history.  To deal with the adaptive length of history interactions, for each batch we find the interaction with the longest history, and zero-pad question-answer pairs with shorter history.  The final LSTM-net hidden state is the history representation $ r_H \in \mathbb{R}^{d_H} $, where $ d_H $ is the history embedding dimension.

\begin{figure}
	\centering
	\includegraphics[width=5cm]{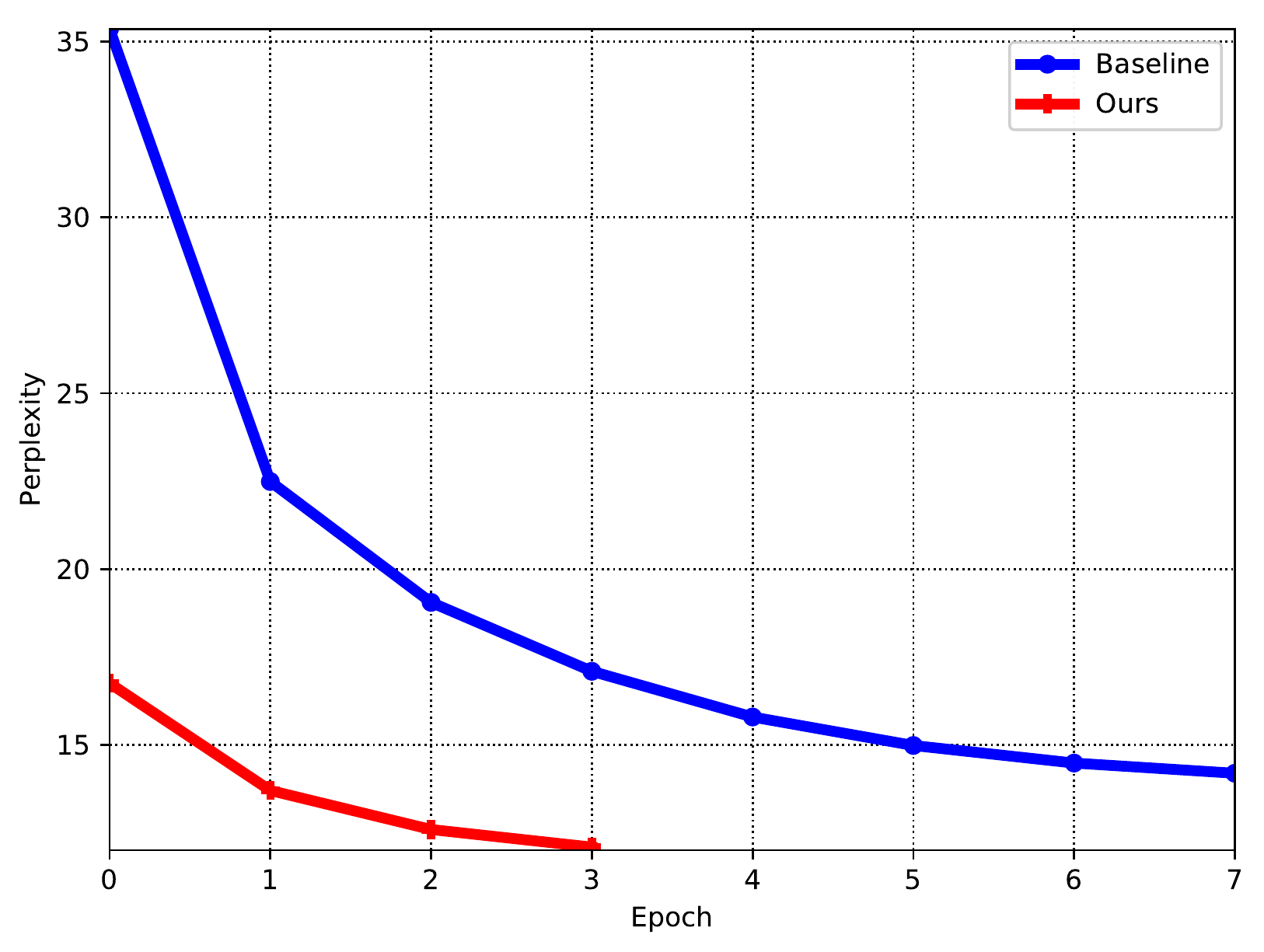}
	\vspace{-0.4cm}
	\caption{\small Perplexity values for our model \vs baseline~\cite{HoriARXIV2018}.}
	\label{fig:train-graph}
	\vspace{-0.5cm}
\end{figure}

\section{Results}
In the following we evaluate the discussed baseline on the 
 Audio Visual Scene-Aware Dialog (AVSD) dataset. We follow the proposed protocol and assess the generated answers to a user question  given a dialog context~\cite{AlamriARXIV2018,HoriARXIV2018}. This context consists of a dialog history (previous questions and answers) in addition to video and audio information about the scene. Our code is publicly available\footnote{https://github.com/idansc/simple-avsd}. 


\subsection{AVSD v0.1 Dataset}
The AVSD dataset consists of annotated conversations about short videos. The dataset contains 9,848 videos taken from CHARADES, a multi-action dataset with 157 action categories~\cite{sigurdsson2016hollywood}. Each dialog is obtained from two Amazon Mechanical Turk (AMT) workers, who discuss about events in a video. One of the workers takes the role of an answerer who had already watched the video. The answerer replies to questions asked by another AMT worker, the questioner. 

The questioner was not shown the  whole video but only the first, middle and last frames of the video. The dialog revolves around the events in and other aspects of the video. 
The AVSD v0.1 dataset is split into 7,659 train dialogs, 	1,787 validation and  1,710 test dialogs. Because the test set doesn't currently include ground truth, we follow~\cite{HoriARXIV2018} and evaluate on the `prototype test-set' with 733 dialogs. 
Because the `prototype test-set' is part of the `v0.1 validation-set,' we use the `prototype validation-set' with 732 dialogs, which doesn't overlap with the `prototype test-set.'

\begin{table}[t]
	
	\centering
	\caption{Results for the AVSD dataset for CIDEr, BLEU1, \ldots, BLEU4, ROUGE-L, METEOR. We provide a comparison to the baseline and a detailed ablation study separated into categories and discussed in \secref{sec:insights}. We also report the number of parameters for each baseline. 
	}	
	\vspace{-0.3cm}
	\resizebox{\linewidth}{!}{
		\begin{tabular}{lcccccccc}
			\Xhline{2\arrayrulewidth} 
			Model   & C & B4    & B3  & B2  & B1  & R  & M & P \\ \hline
			baseline\cite{HoriARXIV2018}\footnote{https://github.com/dialogtekgeek/AudioVisualSceneAwareDialog} & 0.766	& 0.084 & 0.117 & 0.173 & 0.273 & 0.291 & 0.117 & 6.15M\\
			\hline\multicolumn{7}{c}{\textbf{basic baselines}}    \\\hline 
			q   & 0.815 & 0.088 & 0.122  & 0.178  & 0.279   & 0.297 & 0.121  & 3.1M \\ 
			q+h & 0.843 & 0.089 & 0.123   & 0.178  & 0.277  & 0.296 & 0.122 & 4.51M \\
			q+h+vgg-spatial & 0.869 & 0.089 & 0.124 & 0.180 &  0.279 & 0.302 &  0.123 & 5.12M \\
			q+h+vgg-spatial+audio & 0.874	& 0.091 & 0.125 & 0.182 & 0.282 & 0.305 &  0.124 & 5.23M \\
			\hline\multicolumn{7}{c}{\textbf{basic baselines+attention}}    \\\hline
			q+att  & 0.849 & 0.090  &  0.124  & 0.179  & 0.278   & 0.298 & 0.121  & 3.35M \\ 
			q+h+att & 0.861 &  0.090 & 0.124   & 0.177  & 0.271   & 0.298 & 0.122  & 4.57M \\  
			q+h+vgg-spatial+att & 0.908 & 0.093 & 0.129 & 0.185 & 0.283 & 0.307 & 0.125 & 7.4M \\
			\hline\multicolumn{7}{c}{\textbf{attention-model}}    \\\hline
			w/o-cross-data-evidence & 0.896 & 0.095  &  0.131 & 0.190  & 0.292   & 0.309 & \textbf{0.128} & 7.5M \\ 
			w/o-local-evidence & 0.917 &  0.096 & 0.132   & 0.191  & 0.293   &  0.309 & \textbf{0.128} & 8.35M \\  
			w/o-question-prior & 0.906 & 0.096 & 0.132 & 0.190 & 0.292 & 0.309 & 0.127 &8.35M \\
			sharing--weights & 0.923 & \textbf{0.097 }& \textbf{0.133 }& \textbf{0.191} & \textbf{0.293} & 0.309 &  0.127 & 6.18M  \\
			\hline\multicolumn{7}{c}{\textbf{video-fusion}}    \\\hline
			
			temporal-attention & 0.877 & 0.091 & 0.126  & 0.182  & 0.281   & 0.302 & 0.124 &  8.4M \\ 
			summation & 0.890 & 0.093 & 0.128  & 0.183  & 0.283   & 0.303 & 0.124 & 7.35M \\
			weighted-summation & 0.876 & 0.094 & 0.130  & 0.187  & 0.289   & 0.304 & 0.126 & 7.85M \\
			video-audio-lstm	 & 0.865 & 0.076 &  0.101 & 0.141 & 0.210   & 0.286 & 0.108 & 8.35M \\
			
			\hline\multicolumn{7}{c}{\textbf{decoder-input}}    \\\hline 
			q-first-state & 0.704 & 0.078 &  0.110 & 0.163 & 0.257 &  0.279 & 0.112	& 8.35M
			\\ 
			all-first-state & 0.714 & 0.079 &  0.114 & 0.171 & 0.271 & 0.276 & 0.113 & 10.1M \\
			all-concat-decoder-input  & 0.797 & 0.089 & 0.125 & 0.183 & 0.285 & 0.297 & 0.121 & 9.53M \\
			q+h+a-concat-input & 0.857 & 0.090 & 0.123  & 0.177  & 0.274   & 0.298 & 0.121 & 7.72M \\
			\hline\multicolumn{7}{c}{\textbf{i3d-features-\&-spatial-temporal}} \\ \hline
			i3d-rgb-temporal  &  0.886 & 0.094 & 0.130  & 0.188 & 0.289   & 0.306 & 0.126 & 7.23M \\
			i3d-rgb-flow-temporal  & 0.851 & 0.091 &  0.127 & 0.185  & 0.286   & 0.303 & 0.125 & 7.82M  \\  
			i3d-rgb-spatial-10  & 0.928 & \textbf{0.097} & \textbf{0.133}  & 0.190 & 0.290  & 0.310 & 0.127 & 6.58M \\ 
			vgg-spatial-1 & 0.919 & 0.095 & 0.130 & 0.187 & 0.287 & 0.309 & 0.126 & 6.18M \\
			vgg-spatial-16 & 0.903 & 0.093 & 0.128 & 0.186 & 0.287 & 0.307 & 0.127 & 28.88M \\
			\hline\multicolumn{7}{c}{\textbf{initialization}} \\ \hline
			default & 0.877 & 0.090 & 0.123 & 0.178 & 0.274 & 0.300 & 0.121 & 8.35M \\
			xavier & 0.848 & 0.087 & 0.119 & 0.171 & 0.262 & 0.297 & 0.119 & 8.35M \\
			he  & 0.913 & 0.095 &  0.131 & 0.189  & 0.290   & 0.308 & 0.127 & 8.35M  \\
			\hline\multicolumn{7}{c}{\textbf{beam-search hyper-parameters}} \\ \hline
			w/o beam & 0.924 & 0.082 & 0.109 & 0.152 & 0.226 & 0.298 &  0.114 & 8.35M \\
			2-width & 0.934 & 0.094 & 0.128 & 0.183 & 0.279 & 0.311 & 0.126 & 8.35M \\
			4-width  & 0.931 & 0.096 &  0.131 & 0.188  & 0.287   & 0.310 & 0.127 & 8.35M \\ 
			5-width  & 0.926 & 0.096 &  0.132 & 0.188  & 0.289   & 0.309 & 0.127 & 8.35M  \\   
			\Xhline{2\arrayrulewidth}
			\textbf{Ours} 	 & \textbf{ 0.941} & 0.096 &  0.131 & 0.187  & 0.285   & \textbf{0.311} & \textbf{0.128} & 8.35M \\
			\Xhline{2\arrayrulewidth}
	\end{tabular}}	
	\label{tab:ablation}
	\vspace{-0.5cm}
\end{table}

\subsection{Implementation Details}
Our system relies on textual, visual and audio data representations, \ie, $r_\alpha$ for $\alpha \in \{\emph{A}, \emph{Q}, \emph{V}_1, \ldots, \emph{V}_F\}$. For the \textbf{video} representation we randomly sample  $ F =4 $ equally spaced frames, and use the last conv layer of a VGG19 having a dimensions of  $ 7\times7\times512 $. Therefore the visual embedding  dimension is $d_V=512$. After flattening the 2D spatial dimension, we obtain the spatial dimension $ n_V=49 $. 
For \textbf{audio} features we use VGGish that operates on 0.96s log-Mel spectrogram patches extracted from 16kHz audio, and outputs a $ d_A = 128 $ dimensional vector. VGGish inputs overlap by 50\%, therefore an output is provided every 0.48s. 
Dropout parameters before the last FC layer, and the LSTM layers are set to 0.5. 
For the \textbf{question} representation we set the word embedding dimension to $ 128 $. The questions are embedded to $ d_Q = 256 $ dimensional vectors, extracted from the last hidden state of their LSTM-net. The \textbf{history} consists of $T=10$ question-answer pairs, which we denote by $H = (Q,A)_{t\in\{1,\ldots,T\}}$. We use an LSTM-net with a hidden state of $ d_H = 128 $ to encode the history. 

\subsection{Training}
\label{sec:train}
We use a cross-entropy loss on the probabilities, $ p(y_i|y_{<i}, x) $ to train the answer generator, the attention and the embedding layers  jointly end-to-end. The total amount of trainable parameters are 8,359,107. We use the Adam optimizer~\cite{kingma2014adam} with a learning rate of 0.001 and a batch size of 64. During training after each epoch we evaluate our performance on the validation set using a perplexity metric. We stop our training after two consecutive epochs with no improvement. 

We use a standard machine with an Nvidia Tesla M40 GPU for all  our experiments. 
Training our system takes 4 epochs to converge \vs 9 epochs for the baseline (see \figref{fig:train-graph}). 
Each epoch takes 8 minutes \vs 13 minutes for the baseline. In total, training our model takes approximately 30 minutes. 

\begin{figure*}[t]
	
	\centering
	\includegraphics[width=1\linewidth]{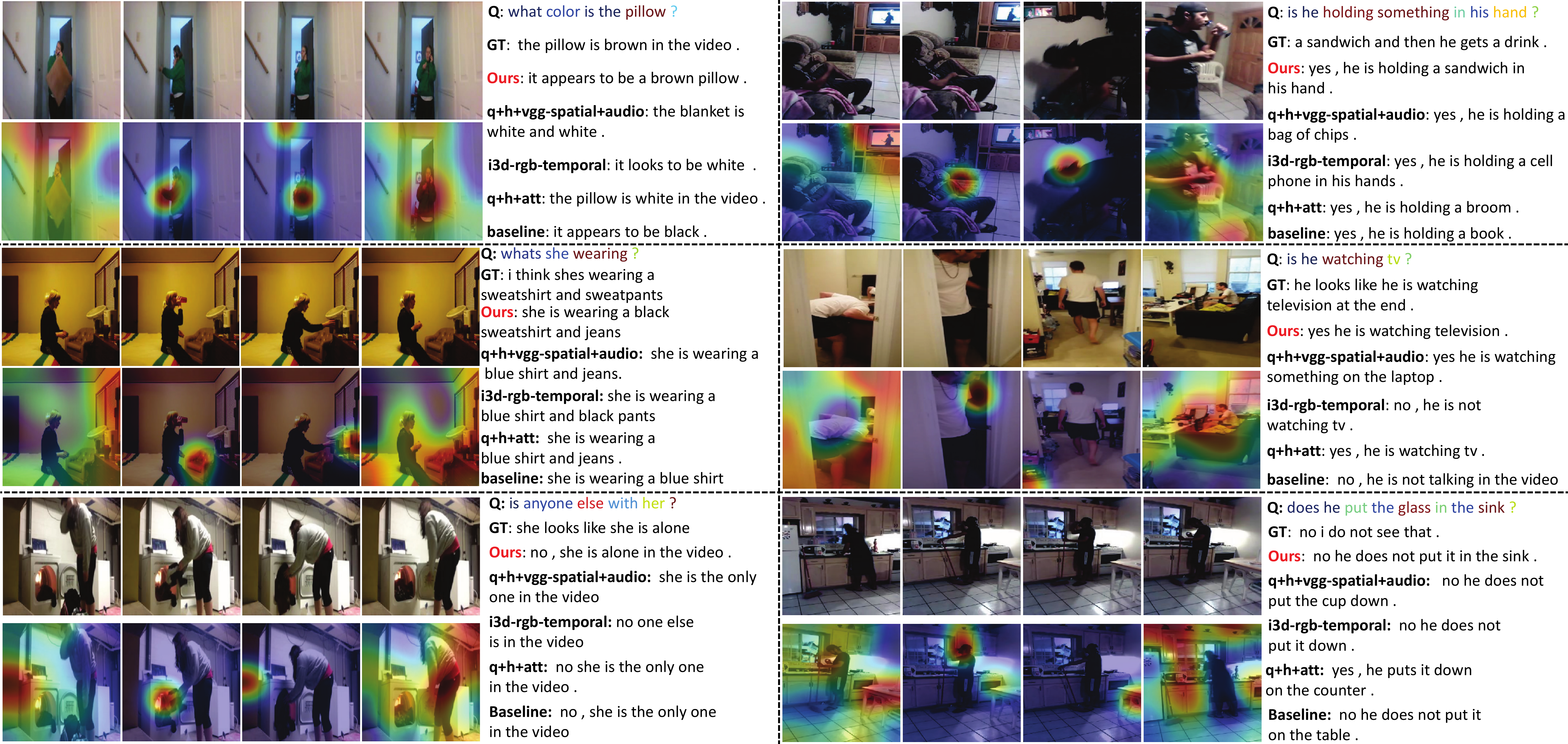}
	\vspace{-0.6cm}
	\caption{\small An illustration of out 4-framed samples from a video along with the relevant attention variables. Our attention treats any frame as different component. This allows the attention module to learn different attention behaviors for different temporal locations. We observe the first and fourth samples are noisier, while the second and third attend to specific interesting locations. Our multimodal attention also generates attention for questions, illustrated over the question via a word heat map. We provide generated answers for different baseline models: q+h+att, is a model with only history and question input; i3d-rgb-temporal is a model with temporal features instead of spatial; q+h+vgg-spatial+audio is a model without attention. We also compare to the generated answer by \cite{HoriARXIV2018}. the ground-truth is denoted by GT, and our final model denoted by Ours.}
	\label{fig:examples}
	\vspace{-0.6cm}
\end{figure*}

\subsection{Performance Evaluation:} 
We evaluate the performance of our system using several metrics. Our prime metric is CIDEr, the Consensus-based Image Description Evaluation, which measures the similarity of a sentence to the consensus~\cite{vedantam2015cider}. We also evaluate our performance on the ROUGE-L metric (Recall Oriented Understudy of Gisting Evaluation). This is a recall-based metric that measures the longest common subsequence of tokens%
~\cite{lin2004rouge}. The METEOR metric is a unigram precision and recall that allows for matchings between candidates and references~\cite{banerjee2005meteor}. We also evaluate our performance using the traditional BLEU score, which measures the effective overlap between a reference sentence and a candidate sentence. We measure the geometric mean of the effective n-gram precision scores, for $n=1,\ldots,4$ and refer to these as BLEU1,$\ldots$, BLEU4.  

\subsection{Quantitative Results and Insights for a Good Baseline}
\label{sec:insights}
We compare to the baseline discussed in~\cite{HoriARXIV2018}.  
In the following we explore the various components of audio-visual dialog systems and present our insights for constructing a simple and effective baseline. These insights cover all aspects of our system: feature embedding, attention, fusion and training techniques. We particularly emphasize the importance of spatial features for AVSD, which we contrast with the action recognition based I3D features. 

\noindent
\textbf{Question Bias and Basic Baselines:}
We revisit the scores published by~\cite{HoriARXIV2018} and assess a basic seq2seq-type baseline, with no attention~\cite{SutskeverNIPS2014}. In this variant, which we call \textbf{q} in \tabref{tab:ablation}, we encode the question using a word embedding (with embedding dimension of $128$) and a 1-layer LSTM-net (with hidden state dimension of $256$ compared to a dimension of 128 in the baseline), without any video or history related features. For decoding,  another 1-layer LSTM-net (with hidden state dimension of $256$ compared to a dimension of 128 in the baseline) is used. 
 Surprisingly, this model alone was able to surpass the current baseline of~\cite{HoriARXIV2018}. Similar results are also reported in~\cite{sanabria2019cmu}. 
 This indicates 
  that there might be bias-problem within the AVSD dataset, no visual information is needed. 
  For instance a common question is ``How many people are in the video?'', but videos in many cases feature  only one person. Another example are questions of the form ``is it indoor?'' which are meaningless since the CHARADES dataset focuses on indoor activities. Another possible explanation for this good result is the encoding of the answer in the  question. For instance, a question ``this person is standing in a kitchen correct?'' is answered with  ``yes he is in the kitchen.'' 
  Moreover, generative evaluation is also more prone to biases, as the evaluation emphasizes correct sentence structure rather than correctness of the answer. Very recently, a discriminative approach was proposed~\cite{alamri2019audio}. The bias problem is not unique to AVSD, and was also discussed for Visual Question Answering~\cite{goyal2017making}.  
 
 To further improve the most basic baseline \textbf{q}, we add more modalities. We use the  fusion  and embedding techniques of the  proposed model 
 but omit attention. Instead of attention, we use a mean over the representation for visual and auditory data sources, and the last hidden state of the LSTM-net is used to represent the question data source. We found that our model can utilize any modality supplement, even without attention. In the `basic baselines+attention' section of \tabref{tab:ablation} we assess versions with attention, which brings us closer to our full model. 


\noindent\textbf{Spatial vs. Temporal Information:} 
\label{sec:spati-temp}
Current methods focus on temporal  models and often na\"ively reduce the spatial dimension~\cite{HoriARXIV2018, wang2016temporal, zhou2017temporal}. 
In contrast, for  closely related visual reasoning tasks, such as visual dialog and visual question answering, it is broadly accepted that spatial attention is necessary. Therefore, it is unlikely that video reasoning is effective when simply reducing the spatial dimension. Indeed, we find better results when reducing the temporal dimension with sampling techniques and employing attention to reduce the spatial dimension. 
In \figref{fig:examples} we observe that a small subset of frames (\eg, 4) is usually enough for an almost complete understanding of the video. 
In the `i3d-features-\&-spatial-temporal' section of \tabref{tab:ablation}, we compare  spatial-based features to  temporal-based ones. The temporal features are  computed on a stack of 16 video frames, and are treated as an  input modality to our attention mechanism. Attention choses the relevant temporal locations. The temporal attended representation was fed to the Aud-Vis LSTM-net along with the audio attended-features. For the i3d-rgb-flow version we also use the I3D model based on optical flow features as an additional data component. This resulted in a drop in performance compared to the spatial-based i3d-features reported in the i3d-rgb-spatial-10 line of \tabref{tab:ablation}. We also test different number of sampled frames. Interestingly, only one frame is already very useful  for AVSD, and too many VGG-frames harm performance. Note that each frame is coupled to an attention-score and treated as a modality, which  explains why too many frames can add noise to the inferred multimodal probability.  

\noindent \textbf{I3D Features \vs VGG:} 
I3D features are widely used as video-based feature extractor (\cf~\cite{carreira2017quo}), discarding the classical image-based features, \eg, VGG.  They are extracted from a model trained on the Kinetics Dataset, a dataset for action recognition, and have been shown to improve many video tasks. We find that while I3D features have repeatedly been shown to improve on action-recognition tasks, they are not as useful in the answer generation task of AVSD. Equipped with VGG features we were able to achieve comparable results to the i3d-rgb-spatial-20 version. The i3d-rgb-spatial features are 4 times bigger (7x7x512 \vs 2x7x7x1024), as well as more complicated to extract. Seeking simplicity, we report  scores with the VGG-based features subsequently.  
This may also indicate a weakness in the dataset, as this solution seems to be sub-optimal for action-related questions (\eg, classifying sequences of actions). 
Not only do we na\"ively  sample temporal frames, but also do we not use I3D features that were extracted from a network trained for action-recognition, yet we achieve good results.

\noindent\textbf{Attention Model:}
\label{sec:abl-atten}
We assess different components of the attention model. See \secref{sec:attention} for details about local evidence and cross data evidence. We found that every component contributes to the model, especially the cross-data component. The cross-data component determines the attention score of an element by considering interactions with other modalities. For instance, a region in the second frame can affect a region in the third frame, or perhaps a word in the question. 

To find the simplest attention module, we also explored the option of grouping together the parameters for all video frames, \ie, $V_{V_1} = \ldots = V_{V_F}$, $L_{V_1} = \ldots = L_{V_F}$, and $R_{V_1} = \ldots = R_{V_F}$, which yields good results  despite 2 million  fewer parameters. 
This version allows to increase the number of processed frames, with no additional memory cost. Those results are reported in the `sharing-weights' line of \tabref{tab:ablation}. 

\noindent\textbf{Multimodal Decoding Fusion:} \label{sec-abl-fuse} 
 We experimented with several variants that reduce $ a_A, a_{V_1}, \ldots, a_{V_F} $. In \tabref{tab:ablation}, section `decoder-input,' we show a version that uses an additional multimodal attention step over the video-related attended vector, called temporal-attention. Another attempt is summation polling of the vectors, and weighted summation with scalers. Instead, we note the sequential information of $ a_{V_1}, \ldots, a_{V_F} $ that naturally calls for the use of an additional LSTM unit, which we call Aud-Vis (see \figref{fig:attention}). 
 We think audio
 is a more general cue while frames have more specific information. Ordering is guided by the intuition that LSTM-based encoding commonly starts with more general information. To verify this intuition, in video-audio-lstm, we performed additional experiments with ordering of $a_{V_1}, \ldots, a_{V_F}, a_A$. 

Next we find a good way to input elements into the answer generation LSTM-net. We first analyze the basic \textbf{q} model. A classic decoder, where encoded \textbf{q} are fed as first hidden state to the LSTM-net is reported in the `q-first-state' row in \tabref{tab:ablation} (decoder-input section). 
This suggest that textual data should be concatenated to the decoder inputs. Concatenating all modalities to the input, which is reported in the `all-concat-input' line in \tabref{tab:ablation} drops the performance, suggesting that a dichotomy of video-related and textual-related features is useful. To incorporate the audio signal, we find it's best to use it as a first state in the Aud-Vis LSTM-net. A version where we concatenated the audio attended vector to  $ a_T $ is referred to as `q+h+a-concat-input+s-first-state.' The model behaves the best when the fused video related features  were used as the initial state $ h_0 $  of the Ans-Generation LSTM-net. Our state-of-the-art model further improves the fusion technique by using the Aud-Vis LSTM-net to generate $h_0$ which captures the temporal information of audio attention $a_A$ and the visual attention $a_{V_1}, \ldots, a_{V_F} $.


\noindent\textbf{Weight Initialization:} An important aspect is the initialization of the deep net parameters. We observed a significant improvement using Kaiming normal initialization or Xavier initialization  for all LSTM models~\cite{HeICCV2015Init,XavierAISTATS2010}. 

\noindent\textbf{Beam Search Width:} In an attempt to improve the overall evaluation time, we  experimented with different  beam width. We found  that although beam search is useful for generation, a width of $2$ achieves almost as good results. Our version use 3-width beam search.

\subsection{Qualitative Results}

In \figref{fig:examples}, we show several examples of generated answers of five models, our final model, a version without any attention (q+h+vgg-spatial+audio), a version with temporal I3D features (i3d-rgb-temporal), a version with only textual modalities (q+h+att), and the baseline~\cite{HoriARXIV2018}. The ground-truth is referred to via GT. Additionally, we take advantage of the interpretability of attention modules to also illustrate the attention probabilities of our final modal on 5 different modalities, \ie, our 4-frames, and the question. First, we observe an interesting behavior of our attention model: each sampled frame is attended a differently, which captures different features from different frames. 
The first and fourth frames are noisier and  extract general concepts, while the second and third  capture unique  aspects of the video, \eg, a person, a couch. This behavior can be associated with the temporal aspect of the frames. Meaning it is more important to capture general aspects at the end and at the beginning, but in the middle we reveal the important specific concepts. Additionally, the question attention attends to the informative words. Our generated answers are usually more aware of the scene, and less prone to bias. For instance,  in the first row, the question is ``what color is the pillow?.'' We observe our model to be able to answer the correct color, while all other model variants answer with white, the most-common color of a pillow. In another question ``whats she is wearing,'' our model was the only one to relate to her black sweatshirt.


\section{Conclusion}
\label{sec:conc}
We propose a simple baseline  for  Audio-Visual Scene-Aware Dialog that surpasses current techniques by 20\% on the CIDEr metric. Pioneering on this task, we carefully evaluated our approach. We hope our analysis can bridge the gap between video-reasoning and image-reasoning. 

\noindent\textbf{Acknowledgments:} This research was supported in part by The Israel Science Foundation (grant
No. 948/15), and by NSF under
Grant No.\ 1718221, Samsung, and 3M. 

{\small
	\bibliographystyle{plain}
	\bibliography{paperrefs}

\begin{thebibliography}{10}

\bibitem{alamri2019audio}
H.~Alamri, V.~Cartillier, A.~Das, J.~Wang, S.~Lee, P.~Anderson, I.~Essa,
  D.~Parikh, D.~Batra, A.~Cherian, T.~K. Marks, and C.~Hori.
\newblock Audio-visual scene-aware dialog.
\newblock {\em arXiv preprint arXiv:1901.09107}, 2019.

\bibitem{AlamriARXIV2018}
H.~Alamri, V.~Cartillier, R.~G. Lopes, A.~Das, J.~Wang, I.~Essa, D.~Batra,
  D.~Parikh, A.~Cherian, T.~K. Marks, and C.~Hori.
\newblock {Audio Visual Scene-Aware Dialog (AVSD) Challenge at DSTC7}.
\newblock In {\em https://arxiv.org/abs/1806.00525}, 2018.

\bibitem{AndreasCVPR2016}
J.~Andreas, M.~Rohrbach, T.~Darrell, and D.~Klein.
\newblock {Deep compositional question answering with neural module networks}.
\newblock In {\em Proc. CVPR}, 2016.

\bibitem{AnejaCVPR2018}
J.~Aneja, A.~Deshpande, and A.~G. Schwing.
\newblock {Convolutional Image Captioning}.
\newblock In {\em Proc. CVPR}, 2018.

\bibitem{AnatolICCV2015}
S.~Antol, A.~Agrawal, J.~Lu, M.~Mitchell, D.~Batra, C.~L. Zitnick, and
  D.~Parikh.
\newblock {VQA: Visual question answering}.
\newblock In {\em Proc. ICCV}, 2015.

\bibitem{aytar2016soundnet}
Y.~Aytar, C.~Vondrick, and A.~Torralba.
\newblock {Soundnet: Learning sound representations from unlabeled video}.
\newblock In {\em Proc. NIPS}, 2016.

\bibitem{banerjee2005meteor}
S.~Banerjee and A.~Lavie.
\newblock Meteor: An automatic metric for mt evaluation with improved
  correlation with human judgments.
\newblock In {\em Proceedings of the acl workshop on intrinsic and extrinsic
  evaluation measures for machine translation and/or summarization}, 2005.

\bibitem{BenyounesICCV2017Mutan}
H.~Ben-younes, R.~Cadene, M.~Cord, and N.~Thome.
\newblock Mutan: Multimodal tucker fusion for visual question answering.
\newblock In {\em Proc. ICCV}, 2017.

\bibitem{carreira2017quo}
J.~Carreira and A.~Zisserman.
\newblock Quo vadis, action recognition? a new model and the kinetics dataset.
\newblock In {\em Proc. CVPR}, 2017.

\bibitem{ChatterjeeECCV2018}
M.~Chatterjee and A.~G. Schwing.
\newblock {Diverse and Coherent Paragraph Generation from Images}.
\newblock In {\em Proc. ECCV}, 2018.

\bibitem{ChenCVPR2015}
X.~Chen and C.~L. Zitnick.
\newblock {Mind's eye: A recurrent visual representation for image caption
  generation}.
\newblock In {\em Proc. CVPR}, 2015.

\bibitem{DasARXIV2016}
A.~Das, H.~Agrawal, C.~L. Zitnick, D.~Parikh, and D.~Batra.
\newblock {Human attention in visual question answering: Do humans and deep
  networks look at the same regions?}
\newblock In {\em Proc. EMNLP}, 2016.

\bibitem{visdial}
A.~Das, S.~Kottur, K.~Gupta, A.~Singh, D.~Yadav, J.~M.~F. Moura, D.~Parikh, and
  D.~Batra.
\newblock {V}isual {D}ialog.
\newblock In {\em Proc. CVPR}, 2017.

\bibitem{visdial_rl}
A.~Das, S.~Kottur, J.~M.~F. Moura, S.~Lee, and D.~Batra.
\newblock Learning cooperative visual dialog agents with deep reinforcement
  learning.
\newblock In {\em Proc. ICCV}, 2017.

\bibitem{DeshpandeARXIV2018}
A.~Deshpande, J.~Aneja, L.~Wang, A.~G. Schwing, and D.~A. Forsyth.
\newblock {Diverse and Controllable Image Captioning with Part-of-Speech
  Guidance}.
\newblock In {\em Proc. CVPR}, 2019.

\bibitem{donahue2015long}
J.~Donahue, L.~Anne~Hendricks, S.~Guadarrama, M.~Rohrbach, S.~Venugopalan,
  K.~Saenko, and T.~Darrell.
\newblock Long-term recurrent convolutional networks for visual recognition and
  description.
\newblock In {\em Proc. CVPR}, 2015.

\bibitem{EphratSIGGRAPH2018}
A.~Ephrat, I.~Mosseri, O.~Lang, T.~Dekel, K.~Wilson, A.~Hassidim, W.~T.
  Freeman, and M.~Rubinstein.
\newblock {Looking to Listen at the Cocktail Party: A Speaker-Independent
  Audio-Visual Model for Speech Separation}.
\newblock In {\em Proc. SIGGRAPH}, 2018.

\bibitem{FukuiARXIV2016}
A.~Fukui, D.~H. Park, D.~Yang, A.~Rohrbach, T.~Darrell, and M.~Rohrbach.
\newblock {Multimodal compact bilinear pooling for visual question answering
  and visual grounding}.
\newblock In {\em Proc. EMNLP}, 2016.

\bibitem{GaoNIPS2015}
H.~Gao, J.~Mao, J.~Zhou, Z.~Huang, L.~Wang, and W.~Xu.
\newblock {Are you talking to a machine? Dataset and Methods for Multilingual
  Image Question Answering}.
\newblock In {\em Proc. NIPS}, 2015.

\bibitem{GaoCVPRW2018}
R.~Gao, R.~S. Feris, and K.~Grauman.
\newblock {Learning to Separate Object Sounds by Watching Unlabeled Video}.
\newblock In {\em Proc. CVPR Workshop}, 2018.

\bibitem{XavierAISTATS2010}
X.~Glorot and Y.~Bengio.
\newblock Understanding the difficulty of training deep feedforward neural
  networks.
\newblock In {\em Proc. AISTATS}, 2010.

\bibitem{goyal2017making}
Y.~Goyal, T.~Khot, D.~Summers-Stay, D.~Batra, and D.~Parikh.
\newblock Making the v in vqa matter: Elevating the role of image understanding
  in visual question answering.
\newblock In {\em Proc. CVPR}, 2017.

\bibitem{HeICCV2015Init}
K.~He, X.~Zhang, S.~Ren, and J.~Sun.
\newblock Delving deep into rectifiers: Surpassing human-level performance on
  imagenet classification.
\newblock In {\em Proc. ICCV}, 2015.

\bibitem{hershey2017cnn}
S.~Hershey, S.~Chaudhuri, D.~P.~W. Ellis, J.~F. Gemmeke, A.~Jansen, R.~C.
  Moore, M.~Plakal, D.~Platt, R.~A. Saurous, B.~Seybold, M.~Slaney, R.~J.
  Weiss, and K.~Wilson.
\newblock Cnn architectures for large-scale audio classification.
\newblock In {\em Proc. ICASSP}, 2017.

\bibitem{HoriARXIV2018}
C.~Hori, H.~Alamri, J.~Wang, G.~Wichern, T.~Hori, A.~Cherian, T.~K. Marks,
  V.~Cartillier, R.~G. Lopes, A.~Das, I.~Essa, D.~Batra, and D.~Parikh.
\newblock {End-to-End Audio Visual Scene-Aware Dialog using Multimodal
  Attention-Based Video Features}.
\newblock In {\em https://arxiv.org/abs/1806.08409}, 2018.

\bibitem{JabriARXIV2016}
A.~Jabri, A.~Joulin, and L.~van~der Maaten.
\newblock {Revisiting Visual Question Answering Baselines}.
\newblock In {\em Proc. ECCV}, 2016.

\bibitem{JainCVPR2018}
U.~Jain, S.~Lazebnik, and A.~G. Schwing.
\newblock {Two can play this Game: Visual Dialog with Discriminative Question
  Generation and Answering}.
\newblock In {\em Proc. CVPR}, 2018.

\bibitem{JainCVPR2017}
U.~Jain$^\ast$, Z.~Zhang$^\ast$, and A.~G. Schwing.
\newblock {Creativity: Generating Diverse Questions using Variational
  Autoencoders}.
\newblock In {\em Proc. CVPR}, 2017.
\newblock $^\ast$ equal contribution.

\bibitem{JohnsonCVPR2017Clevr}
J.~Johnson, B.~Hariharan, L.~van~der Maaten, L.~Fei-Fei, C.~L. Zitnick, and
  R.~Girshick.
\newblock Clevr: A diagnostic dataset for compositional language and elementary
  visual reasoning.
\newblock In {\em Proc. CVPR}, 2017.

\bibitem{johnson2016densecap}
J.~Johnson, A.~Karpathy, and L.~Fei-Fei.
\newblock Densecap: Fully convolutional localization networks for dense
  captioning.
\newblock In {\em Proc. CVPR}, 2016.

\bibitem{KarpathyCVPR2015}
A.~Karpathy and L.~Fei-Fei.
\newblock {Deep visual-semantic alignments for generating image descriptions}.
\newblock In {\em Proc. CVPR}, 2015.

\bibitem{KarpathyCVPR2014}
A.~Karpathy, G.~Toderici, S.~Shetty, T.~Leung, R.~Sukthankar, and L.~Fei-Fei.
\newblock {Large-Scale Video Classification with Convolutional Neural
  Networks}.
\newblock In {\em Proc. CVPR}, 2014.

\bibitem{kim2016hadamard}
J.-H. Kim, K.-W. On, W.~Lim, J.~Kim, J.-W. Ha, and B.-T. Zhang.
\newblock Hadamard product for low-rank bilinear pooling.
\newblock {\em arXiv preprint arXiv:1610.04325}, 2016.

\bibitem{kingma2014adam}
D.~P. Kingma and J.~Ba.
\newblock Adam: A method for stochastic optimization.
\newblock {\em arXiv preprint 2014}.

\bibitem{kottur2018visual}
S.~Kottur, J.~M.~F. Moura, D.~Parikh, D.~Batra, and M.~Rohrbach.
\newblock Visual coreference resolution in visual dialog using neural module
  networks.
\newblock In {\em Proc. ECCV}, 2018.

\bibitem{LiARXIV2017DualVQAVQG}
Y.~Li, N.~Duan, B.~Zhou, X.~Chu, W.~Ouyang, and X.~Wang.
\newblock {Visual Question Generation as Dual Task of Visual Question
  Answering}.
\newblock In {\em https://arxiv.org/abs/1709.07192}, 2017.

\bibitem{li2018videolstm}
Z.~Li, K.~Gavrilyuk, E.~Gavves, M.~Jain, and C.~G.~M. Snoek.
\newblock Videolstm convolves, attends and flows for action recognition.
\newblock {\em CVIU}, 2018.

\bibitem{lin2004rouge}
C.-Y. Lin.
\newblock Rouge: A package for automatic evaluation of summaries.
\newblock {\em Text Summarization Branches Out}, 2004.

\bibitem{lin2014microsoft}
T.-Y. Lin, M.~Maire, S.~Belongie, J.~Hays, P.~Perona, D.~Ramanan,
  P.~Doll{\'a}r, and C.~L. Zitnick.
\newblock Microsoft coco: Common objects in context.
\newblock In {\em Proc. ECCV}, 2014.

\bibitem{lu2017best}
J.~Lu, A.~Kannan, J.~Yang, D.~Parikh, and D.~Batra.
\newblock Best of both worlds: Transferring knowledge from discriminative
  learning to a generative visual dialog model.
\newblock In {\em Proc. NIPS}, 2017.

\bibitem{lu2016hierarchical}
J.~Lu, J.~Yang, D.~Batra, and D.~Parikh.
\newblock Hierarchical question-image co-attention for visual question
  answering.
\newblock In {\em Proc. NIPS}, 2016.

\bibitem{MaARXIV2015}
L.~Ma, Z.~Lu, and H.~Li.
\newblock {Learning to answer questions from image using convolutional neural
  network}.
\newblock In {\em Proc. AAAI}, 2016.

\bibitem{MalinowskiNIPS2014}
M.~Malinowski and M.~Fritz.
\newblock {A Multi-World Approach to Question Answering about Real-World Scenes
  based on Uncertain Input}.
\newblock In {\em Proc. NIPS}, 2014.

\bibitem{MalinowskiICCV2015}
M.~Malinowski, M.~Rohrbach, and M.~Fritz.
\newblock {Ask your neurons: A neural-based approach to answering questions
  about images}.
\newblock In {\em Proc. ICCV}, 2015.

\bibitem{MaoARXIV2014}
J.~Mao, W.~Xu, Y.~Yang, J.~Wang, Z.~Huang, and A.~Yuille.
\newblock {Deep Captioning with Multimodal Recurrent Neural Networks (m-rnn)}.
\newblock In {\em Proc. ICLR}, 2015.

\bibitem{MassicetiARXIV2018}
D.~Massiceti, N.~Siddharth, P.~K. Dokania, and P.~H.~S. Torr.
\newblock {FlipDial: A Generative Model for Two-Way Visual Dialogue}.
\newblock In {\em https://arxiv.org/abs/1802.03803}, 2018.

\bibitem{mostafazadeh2017image}
N.~Mostafazadeh, D.~Brockett, B.~Dolan, M.~Galley, J.~Gao, G.~P. Spithourakis,
  and L.~Vanderwende.
\newblock Image-grounded conversations: Multimodal context for natural question
  and response generation.
\newblock {\em arXiv preprint arXiv:1701.08251}, 2017.

\bibitem{VQG}
N.~Mostafazadeh, I.~Misra, J.~Devlin, M.~Mitchell, X.~He, and L.~Vanderwende.
\newblock Generating natural questions about an image.
\newblock In {\em Proc. ACL}, 2016.

\bibitem{NarasimhanNIPS2018}
M.~Narasimhan, S.~Lazebnik, and A.~G. Schwing.
\newblock {Out of the Box: Reasoning with Graph Convolution Nets for Factual
  Visual Question Answering}.
\newblock In {\em Proc. NIPS}, 2018.

\bibitem{NarasimhanECCV2018}
M.~Narasimhan and A.~G. Schwing.
\newblock {Straight to the Facts: Learning Knowledge Base Retrieval for Factual
  Visual Question Answering}.
\newblock In {\em Proc. ECCV}, 2018.

\bibitem{OwensECCV2018}
A.~Owens and A.~A. Efros.
\newblock {Audio-Visual Scene Analysis with Self-Supervised Multisensory
  Features}.
\newblock In {\em Proc. ECCV}, 2018.

\bibitem{OwensIJCV2018}
A.~Owens, J.~Wu, J.~H. McDermott, W.~T. Freeman, and A.~Torralba.
\newblock {Learning Sight from Sound: Ambient Sound Provides Supervision for
  Visual Learning}.
\newblock {\em IJCV}, 2018.

\bibitem{Perazzi2016}
F.~Perazzi, J.~Pont-Tuset, B.~McWilliams, L.~{Van Gool}, M.~Gross, and
  A.~Sorkine-Hornung.
\newblock A benchmark dataset and evaluation methodology for video object
  segmentation.
\newblock In {\em Proc. CVPR}, 2016.

\bibitem{rennips2015}
M.~Ren, R.~Kiros, and R.~Zemel.
\newblock {Exploring models and data for image question answering}.
\newblock In {\em Proc. NIPS}, 2015.

\bibitem{sanabria2019cmu}
R.~Sanabria, S.~Palaskar, and F.~Metze.
\newblock Cmu sinbadas submission for the dstc7 avsd challenge.
\newblock In {\em DSTC7 at AAAI2019 workshop}, 2019.

\bibitem{santoro2017simple}
A.~Santoro, D.~Raposo, D.~G. Barrett, M.~Malinowski, R.~Pascanu, P.~Battaglia,
  and T.~Lillicrap.
\newblock A simple neural network module for relational reasoning.
\newblock In {\em Proc. NIPS}, 2017.

\bibitem{SchwartzNIPS2017}
I.~Schwartz, A.~G. Schwing, and T.~Hazan.
\newblock {High-Order Attention Models for Visual Question Answering}.
\newblock In {\em Proc. NIPS}, 2017.

\bibitem{SchwartzCVPR2019}
I.~Schwartz, S.~Yu, T.~Hazan, and A.~G. Schwing.
\newblock {Factor Graph Attention}.
\newblock In {\em Proc. CVPR}, 2019.

\bibitem{ShihCVPR2016}
K.~J. Shih, S.~Singh, and D.~Hoiem.
\newblock {Where to look: Focus regions for visual question answering}.
\newblock In {\em Proc. CVPR}, 2016.

\bibitem{ShlizermanCVPR2018}
E.~Shlizerman, L.~Dery, H.~Schoen, and I.~Kemelmacher-Shlizerman.
\newblock {Audio to Body Dynamics}.
\newblock In {\em Proc. CVPR}, 2018.

\bibitem{sigurdsson2016hollywood}
G.~A. Sigurdsson, G.~Varol, X.~Wang, A.~Farhadi, I.~Laptev, and A.~Gupta.
\newblock Hollywood in homes: Crowdsourcing data collection for activity
  understanding.
\newblock In {\em Proc. ECCV}, 2016.

\bibitem{SimonyanNIPS2014}
K.~Simonyan and A.~Zisserman.
\newblock {Two-Stream Convolutional Networks for Action Recognition in Videos}.
\newblock In {\em Proc. NIPS}, 2014.

\bibitem{Simonyan14c}
K.~Simonyan and A.~Zisserman.
\newblock Very deep convolutional networks for large-scale image recognition.
\newblock In {\em Proc. ICLR}, 2015.

\bibitem{SutskeverNIPS2014}
I.~Sutskever, O.~Vinyals, and Q.~V. Le.
\newblock {Sequence to sequence learning with neural networks}.
\newblock In {\em Proc. NIPS}, 2014.

\bibitem{tran2015learning}
D.~Tran, L.~Bourdev, R.~Fergus, L.~Torresani, and M.~Paluri.
\newblock Learning spatiotemporal features with 3d convolutional networks.
\newblock In {\em Proc. ICCV}, 2015.

\bibitem{vedantam2015cider}
R.~Vedantam, L.-C. Zitnick, and D.~Parikh.
\newblock Cider: Consensus-based image description evaluation.
\newblock In {\em Proc. CVPR}, 2015.

\bibitem{VijayakumarARXIV2016}
A.~K. Vijayakumar, M.~Cogswell, R.~R. Selvaraju, Q.~Sun, S.~Lee, D.~Crandall,
  and D.~Batra.
\newblock {Diverse Beam Search: Decoding Diverse Solutions from Neural Sequence
  Models}.
\newblock In {\em Proc. AAAI}, 2018.

\bibitem{VinyalsCVPR2015}
O.~Vinyals, A.~Toshev, S.~Bengio, and D.~Erhan.
\newblock {Show and tell: A neural image caption generator}.
\newblock In {\em Proc. CVPR}, 2015.

\bibitem{WangNIPS2017}
L.~Wang, A.~G. Schwing, and S.~Lazebnik.
\newblock {Diverse and Accurate Image Description Using a Variational
  Auto-Encoder with an Additive Gaussian Encoding Space}.
\newblock In {\em Proc. NIPS}, 2017.

\bibitem{wang2016temporal}
L.~Wang, Y.~Xiong, Z.~Wang, Y.~Qiao, D.~Lin, X.~Tang, and L.~Van~Gool.
\newblock Temporal segment networks: Towards good practices for deep action
  recognition.
\newblock In {\em Proc. ECCV}. Springer, 2016.

\bibitem{wu2017you}
Q.~Wu, P.~Wang, C.~Shen, I.~Reid, and A.~van~den Hengel.
\newblock Are you talking to me? reasoned visual dialog generation through
  adversarial learning.
\newblock {\em arXiv preprint arXiv:1711.07613}, 2017.

\bibitem{wu2018you}
Q.~Wu, P.~Wang, C.~Shen, I.~Reid, and A.~van~den Hengel.
\newblock Are you talking to me? reasoned visual dialog generation through
  adversarial learning.
\newblock In {\em Proc. CVPR}, 2018.

\bibitem{XiongICML2016}
C.~Xiong, S.~Merity, and R.~Socher.
\newblock {Dynamic memory networks for visual and textual question answering}.
\newblock In {\em Proc. ICML}, 2016.

\bibitem{XuARXIV2015}
H.~Xu and K.~Saenko.
\newblock {Ask, attend and answer: Exploring question-guided spatial attention
  for visual question answering}.
\newblock In {\em Proc. ECCV}, 2016.

\bibitem{show_tell}
K.~Xu, J.~Ba, R.~Kiros, K.~Cho, A.~Courville, R.~Salakhudinov, R.~Zemel, and
  Y.~Bengio.
\newblock {Show, Attend and Tell: Neural Image Caption Generation with Visual
  Attention}.
\newblock In {\em Proc. ICML}, 2015.

\bibitem{YangCVPR2016}
Z.~Yang, X.~He, J.~Gao, L.~Deng, and A.~Smola.
\newblock {Stacked attention networks for image question answering}.
\newblock In {\em Proc. CVPR}, 2016.

\bibitem{you2016image}
Q.~You, H.~Jin, Z.~Wang, C.~Fang, and J.~Luo.
\newblock Image captioning with semantic attention.
\newblock In {\em Proc. CVPR}, 2016.

\bibitem{yu2016video}
H.~Yu, J.~Wang, Z.~Huang, Y.~Yang, and W.~Xu.
\newblock Video paragraph captioning using hierarchical recurrent neural
  networks.
\newblock In {\em Proc. CVPR}, 2016.

\bibitem{ZhangICCV2015}
D.~Zhang and M.~Shah.
\newblock {Human Pose Estimation in Videos}.
\newblock In {\em Proc. ICCV}, 2015.

\bibitem{zhou2017temporal}
B.~Zhou, A.~Andonian, and A.~Torralba.
\newblock Temporal relational reasoning in videos.
\newblock {\em arXiv preprint arXiv:1711.08496}, 2017.

\bibitem{ZhuCVPR2016}
Y.~Zhu, O.~Groth, M.~Bernstein, and L.~Fei-Fei.
\newblock {Visual7W: Grounded Question Answering in Images}.
\newblock In {\em Proc. CVPR}, 2016.

\end{thebibliography}
}

\end{document}